\documentclass{article}
\usepackage{iclr2026_conference,times}

\usepackage{amsmath,amsfonts,bm}

\def\eqref#1{equation~\ref{#1}}

\def\1{\bm{1}}

\DeclareMathAlphabet{\mathsfit}{\encodingdefault}{\sfdefault}{m}{sl}
\SetMathAlphabet{\mathsfit}{bold}{\encodingdefault}{\sfdefault}{bx}{n}

\usepackage{graphicx}
\usepackage{caption}
\usepackage[table]{xcolor}
\usepackage{booktabs}
\usepackage{multirow}
\usepackage{makecell}
\usepackage{wrapfig}
\usepackage{enumitem}

\usepackage{amssymb}
\usepackage{mathtools}     
\usepackage{amsthm}

\usepackage{tcolorbox}
\usepackage{algorithm2e}
\usepackage{fontawesome}

\definecolor{deepgreen}{rgb}{0,0.5,0}
\definecolor{deepred}{rgb}{0.5,0,0}
\definecolor{citation}{RGB}{10,110,150}

\theoremstyle{plain}

\theoremstyle{definition}

\theoremstyle{remark}

\usepackage{hyperref}
\hypersetup{
    colorlinks=true,
    linkcolor=citation,
    anchorcolor=citation,
    citecolor=citation
}

\newtcolorbox{promptbox}[1][]{
    colback=gray!20!white,
    colbacktitle=white,
    coltitle=black,
    colframe=black!75!black,
    boxrule=0.7pt,
    halign title=center,
    title=\textbf{#1}
}

\title{\includegraphics[height=1.2em]{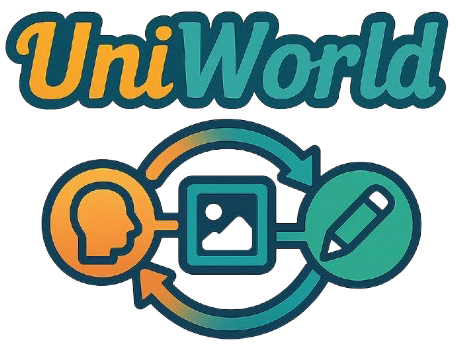} \texttt{Uniworld-V2}: Reinforce Image Editing with Diffusion Negative-aware Finetuning and MLLM Implicit Feedback}

\author{
}

\makeatletter
\newcommand{\eg}{\emph{e.g.}\@ifnextchar.{\!}{\ }}
\makeatother

\iclrfinalcopy 
\begin{document}

\maketitle

\vspace{-5.5em}
\begin{center}
\textbf{UniWorld Team}

\textsuperscript{1}Shenzhen Graduate School, Peking University
\hspace{1.5em}
\textsuperscript{2}Rabbitpre AI \\
\vspace{2pt}
Full author list in Contributions
\end{center}

\begin{table}[ht]
  \centering
  \footnotesize
  \begin{tabular}{@{}l l@{}}
    \faGithub\quad  \href{https://github.com/PKU-YuanGroup/Uniworld}{Github} \quad \quad
    \raisebox{-0.2\height}{\includegraphics[height=1.1em]{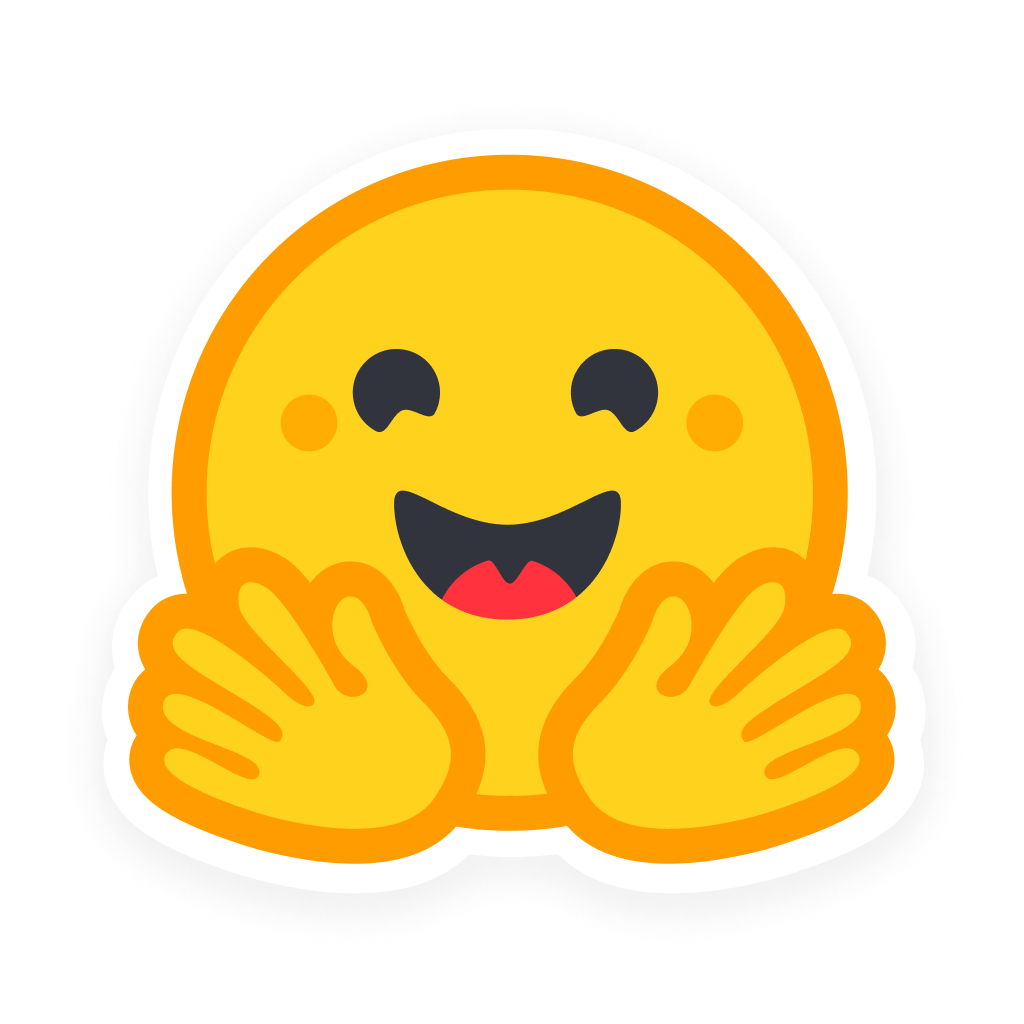}}\quad \href{https://huggingface.co/collections/chestnutlzj/edit-r1-68dc3ecce74f5d37314d59f4}{HuggingFace}  \quad \quad

  \end{tabular}
\end{table}

\renewcommand{\sectionautorefname}{Section}
\renewcommand{\subsectionautorefname}{Section}

\begin{abstract}
Instruction-based image editing has achieved remarkable progress; however, models solely trained via supervised fine-tuning often overfit to annotated patterns, hindering their ability to explore and generalize beyond training distributions.
To this end, we introduce Edit-R1, a novel post-training framework for instruction-based image editing based on policy optimization. 
Specifically, we utilize Diffusion Negative-aware Finetuning (DiffusionNFT), a likelihood-free policy optimization method consistent with the flow matching forward process, thereby enabling the use of higher-order samplers and more efficient training.
Another key challenge here is the absence of a universal reward model, resulting from the diverse nature of editing instructions and tasks. To bridge this gap, we employ a Multimodal Large Language Model (MLLM) as a unified, training-free reward model, leveraging its output logits to provide fine-grained feedback.
Furthermore, we carefully design a low-variance group filtering mechanism to reduce MLLM scoring noise and stabilize optimization.
\texttt{UniWorld-V2}, trained with this framework, achieves \textbf{state-of-the-art} results on the ImgEdit and GEdit-Bench benchmarks, scoring 4.49 and 7.83, respectively. 
Crucially, our framework is model-agnostic, delivering substantial performance gains when applied to diverse base models like Qwen-Image-Edit and FLUX-Kontext, demonstrating its wide applicability.
Code and models are publicly available to support further research.
\end{abstract}

\begin{figure}[ht]
    \centering
    \includegraphics[width=1.0\linewidth]{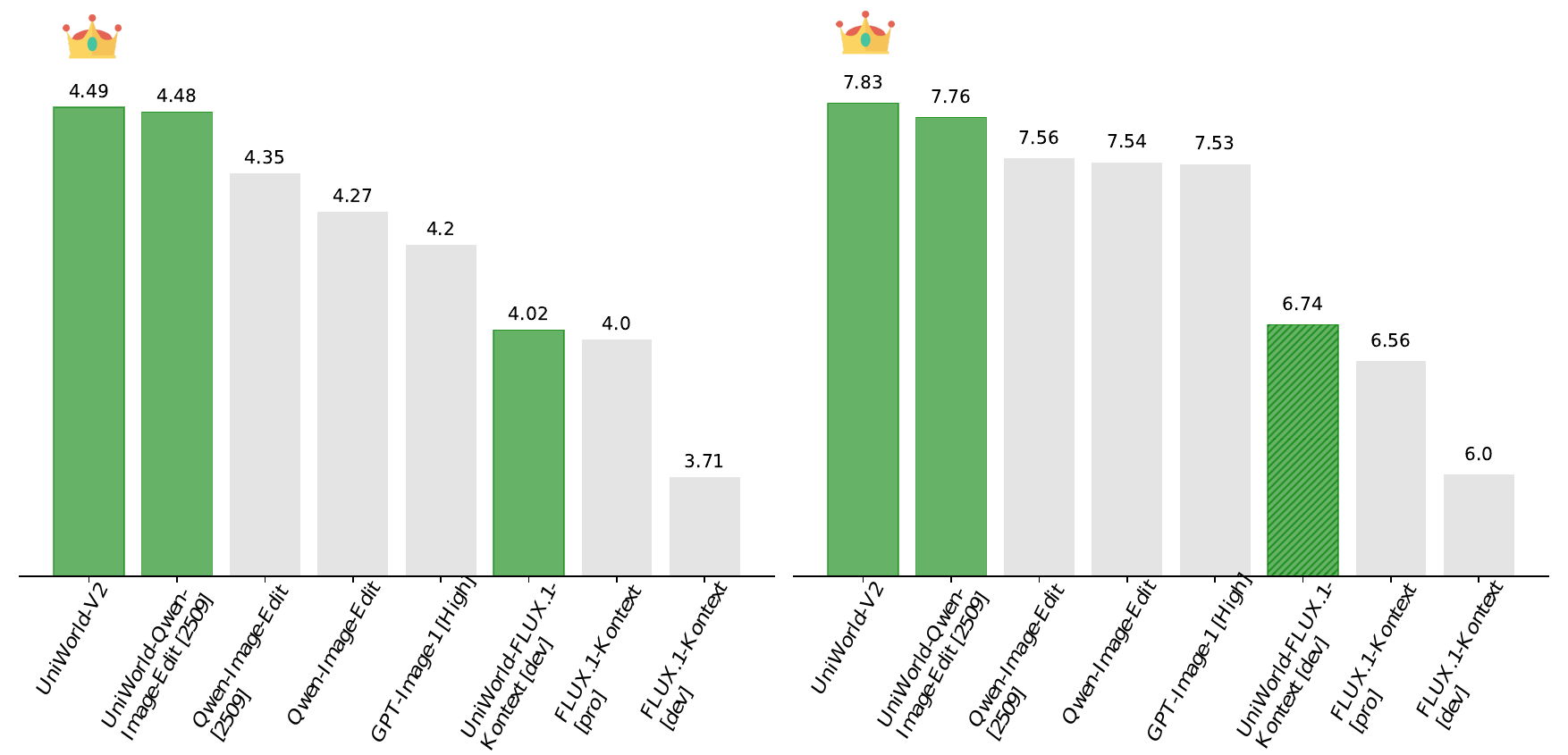} 
    \caption{
        On ImgEdit~\citep{ye2025imgedit} and GEdit-Bench~\citep{liu2025step1x} leaderboards, our method achieves \textbf{state-of-the-art} performance.
    }
    \label{fig_bench}
\end{figure}

\begin{figure}
    \hspace{-2em}
    \vspace{-3em}
    \includegraphics[width=1.1\linewidth]{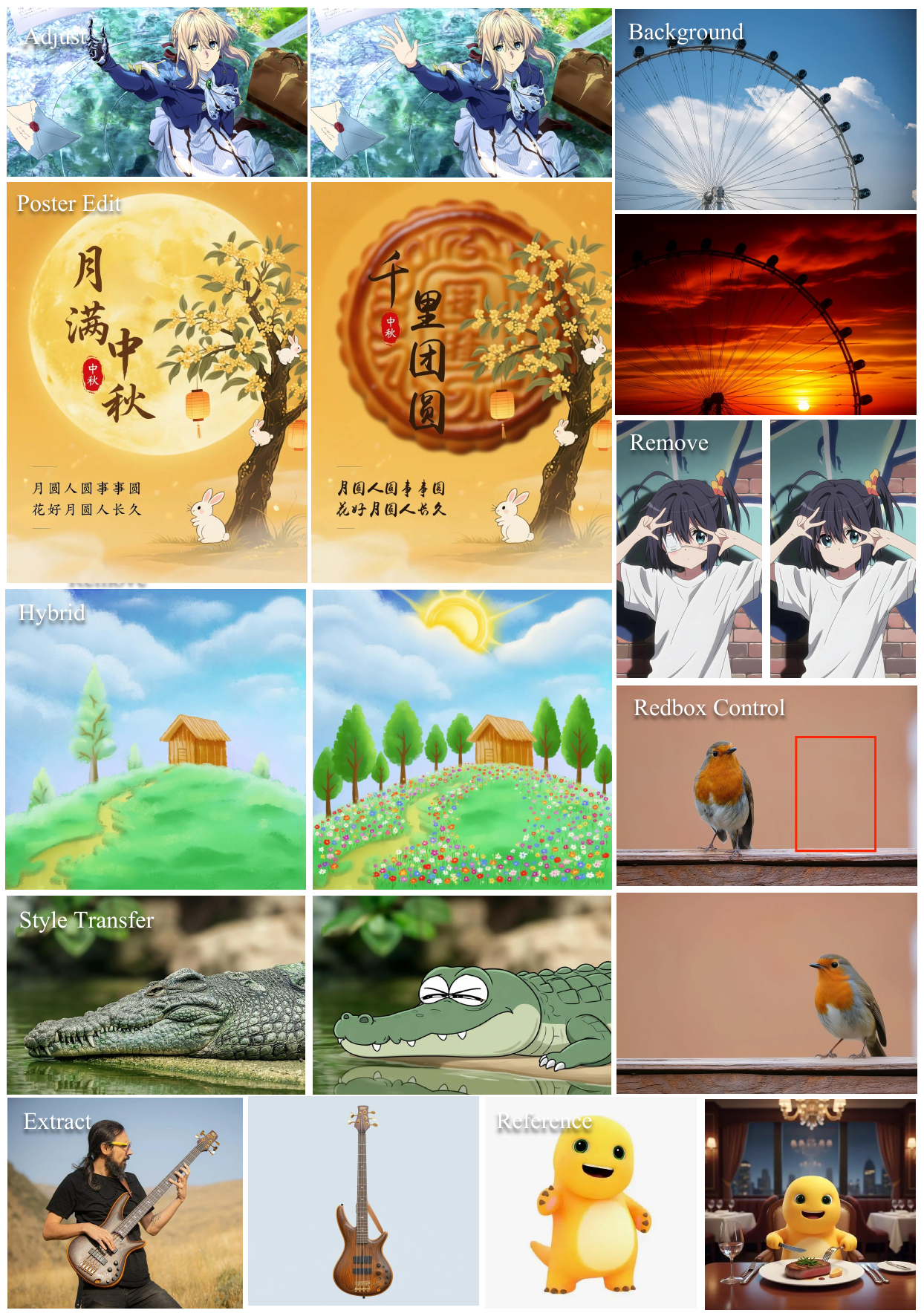}
    \label{fig_head}
\end{figure}

\section{Introduction}
Recent advances in diffusion models~\citep{song2020score,rombach2022high,lipman2022flow} have significantly improved Text-to-Image (T2I) generation~\citep{li2024hunyuan,lin2025uniworld,wu2025qwen,labs2025flux1kontextflowmatching,yan2025can}, enabling high-quality and diverse image synthesis. Building on this, diffusion models are increasingly being extended to image editing, which requires precise instruction-following and fine-grained control while preserving unedited content.

Image editing methods include workflow-based~\citep{rombach2022high,zhang2023adding,ye2023ip} and instruction-based approaches. While recent instruction-based models offer the convenience of unified, prompt-driven editing within a single architecture~\citep{brooks2023instructpix2pix,wu2025qwen,labs2025flux1kontextflowmatching,lin2025uniworld,deng2025emerging,liu2025step1x,openai_image_api}, they often struggle with generalization and controllability. These shortcomings stem from the inherent limitations of the Supervised Fine-Tuning (SFT) paradigm. The SFT objective tends to shortcut learning, causing models to ignore complex instructions and revert to merely reconstructing the input~\citep{labs2025flux1kontextflowmatching, wu2025qwen, liu2025step1x}. Furthermore, its reliance on large-scale but less diverse datasets makes models prone to overfitting, compromising their instruction fidelity across diverse tasks. To overcome these issues and better align models with human intent, post-training alignment via Reinforcement Learning (RL) has emerged as a promising direction.

Motivated by recent advances in applying RL to language models~\citep{shao2024deepseekmath,yu2025rlprextrapolatingrlvrgeneral}, several works~\citep{liu2025flow,xue2025dancegrpo} have explored integrating RL with diffusion models. However, recent studies~\citep{zheng2025diffusionnft, xue2025advantage} highlight that policy optimization methods employing likelihood estimation can introduce systematic bias and limit solver flexibility. Moreover, these methods~\citep{liu2025flow,xue2025dancegrpo} relying on first-order SDE samplers force a trade-off between trajectory diversity and generation quality. These challenges are particularly acute in image editing, where both high-fidelity generation and diverse exploration are crucial for achieving satisfactory results.

Beyond policy optimization, the success of RL depends on a high-quality reward signal. MLLMs are well-suited for the subjective evaluation required in image editing, offering assessments aligned with human intent. Existing MLLM scoring methods can be categorized as logit-based~\citep{wu2024q,zhang2024q,gong2025onereward,li2025generalist,xu2023imagereward}, which use token distribution statistics for interpretability, and sampling-based~\citep{wang2025unified,xu2024visionreward, luo2025editscore, wu2025editreward}, which extract scores from generated outputs. 
While some approaches~\citep{wang2025unified} use Chain-of-Thought (CoT) to improve reward accuracy, they can introduce exposure bias~\citep{huang2025survey} and high computational costs, leading to a skewed reward distribution under logit-based scoring. 
Alternatively, fine-tuning improves domain-specific scoring, yet demanding carefully designed datasets, which are necessary to avoid bias and catastrophic forgetting~\citep{huang2025survey} and costly for diverse editing tasks. Therefore, an ideal reward mechanism should leverage the powerful visual understanding priors of MLLMs without costly fine-tuning or unreliable complex reasoning.

We advance post-training for editing models along two main directions: (1) adopting a more effective policy optimization method and (2) leveraging pre-trained MLLMs for reliable, low-cost, and low-hallucination reward signals. In this work, we propose an efficient post-training framework, Edit-R1, which provides an integrated solution to both challenges. First, we adopt DiffusionNFT~\citep{zheng2025diffusionnft} for policy optimization. This method decouples training and sampling, supports black-box solvers, and removes the need for likelihood estimation. Second, we introduce a novel training-free reward model derived from pretrained MLLMs that does not require CoT reasoning. Instead of sampling-based scoring, we apply a logit-based scoring mechanism to compute expected scores directly from token logits, improving interpretability and computational efficiency.

To validate Edit-R1, we apply our method to diverse base models, including UniWorld-V2~\citep{lin2025uniworld}, FLUX.1 Kontext [Dev]~\citep{labs2025flux1kontextflowmatching}, and Qwen-Image-Edit [2509]~\citep{wu2025qwen}. It elevates FLUX.1-Kontext [Dev] surpasses its stronger Pro version and sets a new open source state-of-the-art when applied to Qwen-Image-Edit [2509], achieving scores of 4.48 on ImgEdit and 7.79 on GEdit-Bench.
Moreover, UniWorld-V2 yields state-of-the-art performance of 4.49 and 7.83, respectively, outperforming prominent closed-source models.
These results are shown in~\autoref{fig_bench} and underscore that our model framework consistently unlocks the potential within various models, showcasing strong generalization.

Our main contributions are as follows:
\begin{itemize}[nolistsep, leftmargin=*]
    \item We propose the Edit-R1 framework, which employs DiffusionNFT and a training-free reward model derived from pretrained MLLMs to fine-tune diffusion models for image editing.
    
    \item We validate that our reward signal offers superior alignment with human preferences, providing reliable, low-cost, and low-hallucination reward signals that stabilize training.
    
    \item Experimental results show that our method significantly improves the performance of UniWorld-V2, Qwen-Image-Edit, and FLUX.1-Kontext across diverse editing benchmarks.

\end{itemize}
\section{Methodology}
\subsection{Preliminary}
\textbf{Flow Matching.} Given a data sample $x_0 \sim X_0$ with a corresponding condition $c$ (e.g., a class label or text embedding). from the true distribution and a gaussian noise sample $x_1 \sim X_1$, Rectified Flow ~\citep{liu2022flow,lipman2022flow} defines an interpolation noised sample $x_t$ as,
\begin{equation}
\label{eq:flow_forward}
    x_t = (1-t)x_0 + tx_1,
\end{equation}
where \(t \in [0,1]\). Given $c$ as the text embedding, a neural network $v_\theta(x_t, t, c)$ is trained to approximate the target velocity field $v = x_1 - x_0$ by minimizing the flow matching objective:
\begin{equation}
\label{eq:flow_matching}
    \mathcal{L}_{FM}(\theta) = \mathbb{E}_{t,x_0 \sim X_0,x_1 \sim X_1}[||v - v_{\theta}(x_t, t, c)||_2^2].
\end{equation}
Inference is performed by solving a deterministic ODE for the forward process:
\begin{equation}
\label{eq:forward}
    dx_t = v_{\theta}(x_t, t, c)dt.
\end{equation}

\begin{figure}[t]
    \centering
    \includegraphics[width=\linewidth]{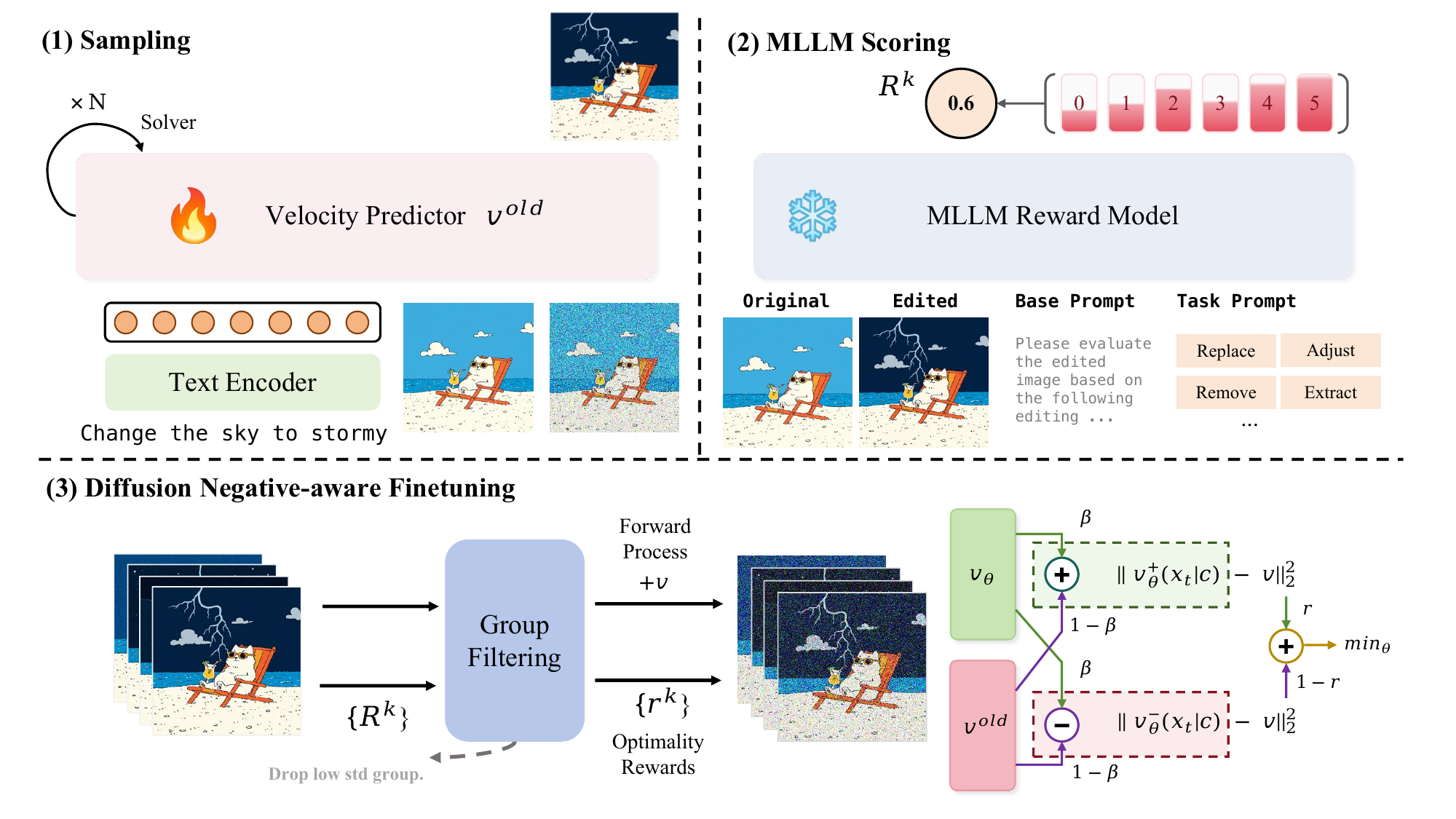} 
    \caption{
        \textbf{Overview of the Edit-R1 pipeline}.
        Our framework consists of three parts: 
        1) We employ the DPM-Solver~\citep{lu2022dpm} to perform a rapid rollout, generating a group of candidate images from the  policy.
        2) We use implicit feedback from MLLM to score the image editing effect and provide rewards. The scoring instructions include both a basic instruction for general editing requirements and a task instruction designed for fine-grained scoring based on the specific editing task type.
        3) We fine-tune the velocity predictor using DiffusionNFT~\citep{zheng2025diffusionnft}, enhanced by \textit{group filtering} method that removes low-variance groups.
    }
    \label{fig_pipeline} 
\end{figure}

\textbf{Diffusion Negative-aware Finetuning (DiffusionNFT).} 
Unlike RL algorithms~\citep{shao2024deepseekmath, liu2025flow, xue2025dancegrpo}, built upon the policy gradient framework, DiffusionNFT~\citep{zheng2025diffusionnft} performs policy optimization directly on the forward diffusion process via the flow matching objective. The method leverages a reward signal $r(x_0, c)$ to define a contrastive loss that steers the model's velocity predictor, $v_\theta$, toward the high-reward policy and away from the low-reward one. The core policy optimization loss is defined as:
\begin{equation}\label{eq:policy_opt}
\mathcal{L}(\theta)
=\mathbb{E}_{c,\pi^{\mathrm{old}}(x_0\mid c),t}\Big[ r\,
\|v^+_\theta(x_t,c,t)-v\|_2^2 + (1-r)\,
\|v^-_\theta(x_t,c,t)-v\|_2^2\Big],
\end{equation}
where $v$ is the target velocity field. The implicit positive and negative policies $v^+_\theta$ and $v^-_\theta$ are combinations of the old policy $v^{old}$ and the training policy $v_\theta$, weighted by a hyperparameter $\beta$:
\begin{align}
v^+_\theta(x_t,c,t)&:=(1-\beta)\,v^{old}(x_t,c,t)+\beta\,v_\theta(x_t,c,t), \\
v^-_\theta(x_t,c,t)&:=(1+\beta)\,v^{old}(x_t,c,t)-\beta\,v_\theta(x_t,c,t).
\end{align}
The optimality probability $r \in [0, 1]$ is transformed from the unconstrained raw reward signal $r^{\text{raw}}$:
\begin{equation}
r(x_0, c) := \frac{1}{2} + \frac{1}{2} \text{clip} \left[ \frac{r^{\text{raw}}(x_0, c) - \mathbb{E}_{\pi^{\mathrm{old}}(\cdot | c)} r^{\text{raw}}(x_0, c)}{Z_c}, -1, 1 \right],
\end{equation}
where $Z_c > 0$ is a normalizing factor, such as the global std of the rewards.

\subsection{Training-Free MLLM Scoring}
\label{mllmscore}
Our approach leverages a pretrained MLLM as a training-free reward model to evaluate the editing accuracy. An editing task is defined by an input sequence $\mathbf{X}=(I_{\text{original}}, I_{\text{edited}}, T_{\text{instruction}})$, containing the original image, the edited image, and the text instruction. The MLLM's response generation is modeled as a sequential token-by-token process. Let $R_{<n} = (r_0, r_1, \dots, r_{n-1})$ . The generation of the next token $r_{n+1}$ is conditioned on the previous tokens:

\begin{align}
    p(r_n | \mathbf{X}, R_{<n}) = softmax(\mathcal{D}(\mathbf{X}, R_{<n})).
\end{align}

Here, $\mathcal{D}$ denotes an MLLM, and its output is the logit vector corresponding to the final token in the sequence. We explore the MLLM's evaluation framework along two dimensions: Chain-of-Thought (CoT) versus non-CoT Scoring, and Sampling-based(discrete) versus logit-based(continuous) Scoring.

\textbf{CoT vs. non-CoT} \quad
This dimension explores whether the MLLM should generate explanatory reasoning before providing the final score. 
In non-CoT Scoring, the MLLM directly produces a score without reasoning, resulting in a response length $n=1$. Conversely, CoT Scoring requires the MLLM to generate a CoT reasoning before giving a score, which leads to a response length $n>1$.

\textbf{Sampling-based vs. logit-based}\quad 
This dimension explores how the MLLM's output is converted into a reward signal. First, the MLLM $\mathcal{D}$ generates a textual response $R$ based on a predefined template.
Sampling-based Scoring extracts explicit numerical scores from $R$ via deterministic rules.
\begin{align}
    s_{\text{dis}}(\mathbf{X}) = \text{Parse}(R).
\end{align}
This method is simple but yields sparse signals, disregarding the model uncertainty when scoring.
logit-based Scoring provides a more fine-grained reward, which is calculated as the expected numerical value of the score tokens:
\begin{align}
    s_{\text{con}}(\mathbf{X}) = \sum_{r \in \mathcal{M}} w(r) \cdot p(r_n=r | \mathbf{X}, R_{<n}),
\label{equ:Es}
\end{align}
where $w(r)$ is the numerical value of the token $r$, and $\mathcal{M}$ denotes the set of tokens used for scoring. This score captures the model's confidence distribution across scores. Then we normalize scores to the range $[0, 1]$:

\begin{align}
    \overline{s}(\mathbf{X}) = \frac{s(\mathbf{X}) - \min_{r \in \mathcal{M}} w(r)}{\max_{r \in \mathcal{M}} w(r) - \min_{r \in \mathcal{M}} w(r)}.
\end{align}

\renewcommand{\sectionautorefname}{Appendix}
\renewcommand{\subsectionautorefname}{Appendix}

In Edit-R1, we utilize the non-CoT and logit-based scoring method and $\mathcal{M}=\{0, 1, 2, 3, 4, 5\}$. Furthermore, we evaluate it against other scoring mechanisms to validate its effectiveness. These include sample-based methods and CoT variants that prompt the MLLM to reason before scoring. Additionally, we benchmark against a pre-trained reward model from existing work~\citep{wang2025unified}. For detailed descriptions of each method, please refer to~\autoref{reward_def}.

\renewcommand{\sectionautorefname}{Section}
\renewcommand{\subsectionautorefname}{Section}

\subsection{Low-STD Group Filtering}

\begin{wrapfigure}{r}{0.4\linewidth}
    \centering
    \vspace{-9pt}
    \includegraphics[width=\linewidth]{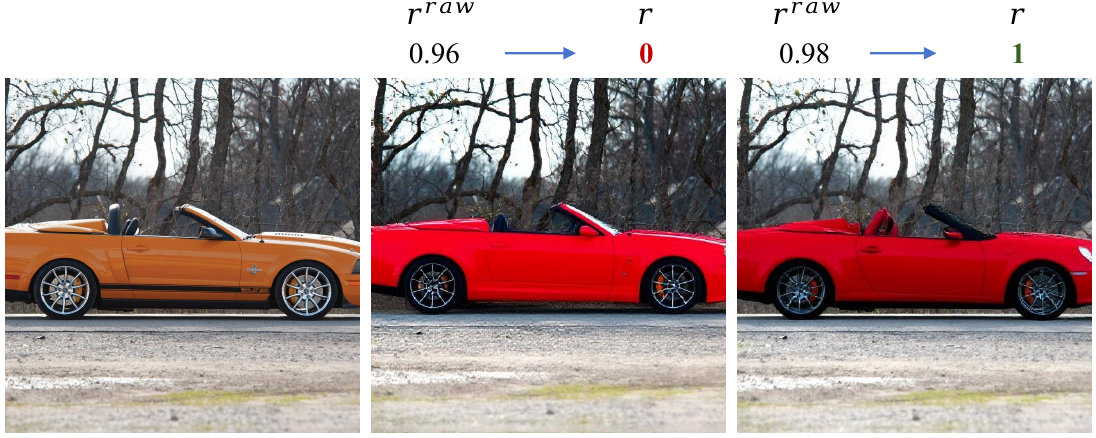} 
    \caption{An example of ``change the car to red'' where samples within a group are typically successful, causing noise amplification after normalization.}
    \label{groupfilter_case} 
    \vspace{-10pt}
\end{wrapfigure}

A potential limitation arises from the normalization operation under conditions of low reward variance. When the probabilities assigned by the MLLM to in-group samples are very similar (\eg, all above 0.95), the minor differences between them cannot reliably indicate a true quality gap. However, in low-variance scenarios, dividing by a standard deviation exaggerates these insignificant scoring differences, as illustrated in \autoref{groupfilter_case}. The resulting reward signal, which reflects noise rather than true quality, can mislead the training process. Filtering out these noisy groups is crucial for maintaining training stability.

Therefore, we aim to filter out groups with high means and low variance in raw rewards. Specifically, we introduce two hyperparameters, $\tau_{\mu}$ and $\tau_{\sigma}$, which represent the thresholds for the mean and variance, respectively. During training, gradients from groups whose mean reward exceeds $\tau_{\mu}$ and whose variance falls below $\tau_{\sigma}$ are discarded and do not contribute to the optimization process.

\subsection{Pipeline of \texttt{Edit-R1}}

To enhance the image editing model, we leverage DiffusionNFT~\citep{zheng2025diffusionnft} and adopt reward signals from an MLLM. This approach makes the reward signal universally applicable to any editing tasks, generating stable rewards from the same distribution for policy optimization while eliminating the reliance on domain-specific reward models. 

As illustrated in~\autoref{fig_pipeline}, the process consists of three main parts: sampling, MLLM scoring, and diffusion negative-aware finetuning, which progressively align the model with the optimal policy.

\textbf{Part 1: Sampling}\quad
Benefiting from the decoupling of policy optimizing and data sampling, DiffusionNFT enables the full utilization of any black-box solvers throughout sampling. Therefore, we specifically employ the DPM-Solver~\citep{lu2022dpm} to perform a rapid rollout for a given source image and an edit instruction, generating a set of $G$ images $\{x_0^i\}_{i=1}^{G}$ sampled from the policy $\pi_{\text{old}}$.

\textbf{Part 2: MLLM scoring}\quad
We evaluate the generated image group $\{x_0^i\}_{i=1}^{G}$ based on implicit feedback from the MLLM to measure alignment with the editing instruction and overall image quality. Conditioned on original image, edited image, and evaluation prompt, the MLLM generates a series of raw reward scores $\{s^i\}_{i=1}^{G}$ for $\{x_0^i\}_{i=1}^{G}$. To facilitate fine-grained scoring, the evaluation prompt is structured into two components: a base prompt, which outlines fundamental editing requirements and instructions, and a task prompt, which is specifically tailored to the type of editing task.

\textbf{Part 3: DiffusionNFT}\quad
The raw MLLM scores $\{s^i\}_{i=1}^{G}$ are transformed into optimality rewards $\{r^i\}_{i=1}^{G}$ through group computation. These rewards are then used to update the policy model $v_{\theta}$ using the DiffusionNFT objective, as defined in~\autoref{eq:policy_opt}. This process guides the model's velocity predictor towards high-reward outcomes while moving it away from low-reward ones, effectively fine-tuning the model to better adhere to user instructions and produce higher-quality edits.
\section{Experiments}

\subsection{Dataset}

\begin{figure}[ht]
    \centering
    \includegraphics[width=\linewidth]{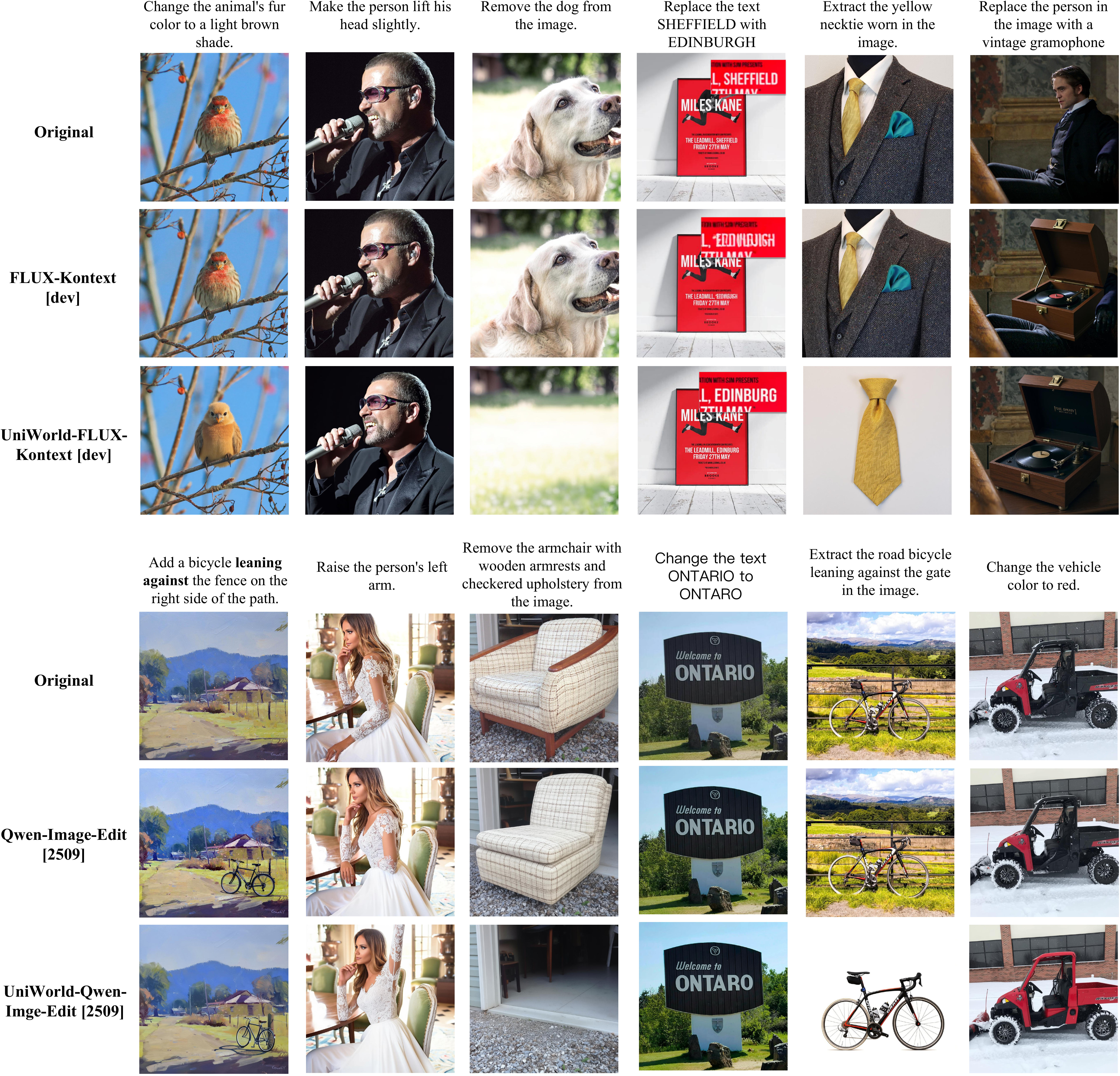} 
    \caption{\textbf{Qualitative comparison of basic editing capabilities before and after policy optimization.}. Basic editing refers to single-step modifications applied to an image.} 
    \label{fig_case} 
\end{figure}

\begin{wrapfigure}{r}{0.5\linewidth}
    \vspace{12pt}
    \centering
    \includegraphics[width=\linewidth]{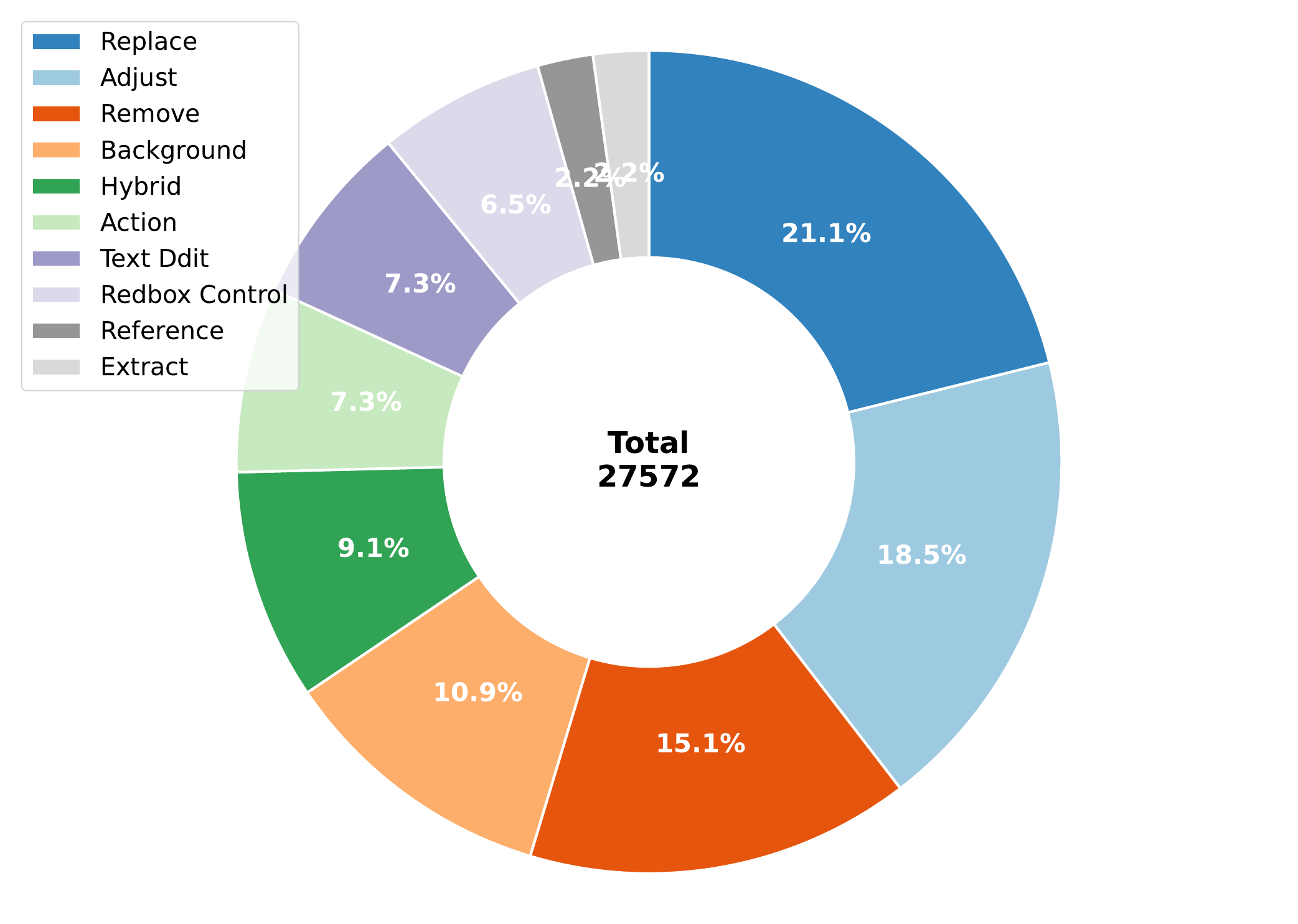} 
    \caption{\textbf{Data Composition Overview}. Our dataset comprises 9 tasks: Replace, Adjust, Remove, Background, Hybrid, Action, Text Edit, Redbox Control, Reference, and Extract.}
    \label{data_bar} 
\end{wrapfigure}

We curate a dataset comprising 27,572 instruction-based editing samples in total (\autoref{data_bar}), which are sourced from LAION~\citep{schuhmann2022laion}, LexArt~\citep{zhao2025lex}, and UniWorld-V1~\citep{lin2025uniworld}. To enhance task diversity, we incorporate additional text-editing and red-box control tasks, resulting in nine distinct task types. Leveraging an online learning paradigm, our method operates solely on the original images and their corresponding editing instructions, which eliminates the need for high-quality edited result images.

For the LAION subset, we utilize the existing object annotations and bounding boxes provided by ImgEdit~\citep{ye2025imgedit}. The preprocessing pipeline includes: 1) Filtering out bounding boxes that are either too small or excessively large. 2) Using the Qwen2.5-VL-32B model to assess the rationality of the editing instructions.
For the Text Edit task, we build upon the LexArt subset by randomly altering characters in words to generate training samples. In the Redbox Control task, we extract a subset from the processed LAION data, draw red bounding boxes around target objects, and generate three types of edit instructions: Adjust, Remove, and Replace.
For the Reference and Extract tasks, we employ high-quality try-on data from UniWorld-V1. Due to the limited diversity in this dataset, we use only 600 samples for each of these two tasks.

\begin{table*}[t]
\centering
\caption{Quantitative comparison results on ImgEdit~\citep{ye2025imgedit}. We use GPT4.1 for evaluation. \textbf{Bold} indicates the best performance.}
\vspace{-0.5em}
\setlength{\tabcolsep}{5pt}
\scalebox{0.63}{
\renewcommand{\arraystretch}{1.1}
\begin{tabular}{l|cccccccccc}
\toprule[1.5pt]
\textbf{Model} & \textbf{Add} & \textbf{Adjust} & \textbf{Extract} & \textbf{Replace} & \textbf{Remove} & \textbf{Background} & \textbf{Style} & \textbf{Hybrid} & \textbf{Action} & \textbf{Overall} $\uparrow$ \\
\midrule
MagicBrush~\citep{zhang2023magicbrush} & 2.84 & 1.58 & 1.51 & 1.97 & 1.58 & 1.75 & 2.38 & 1.62 & 1.22 & 1.90 \\
Instruct-Pix2Pix~\citep{brooks2023instructpix2pix} & 2.45 & 1.83 & 1.44 & 2.01 & 1.50 & 1.44 & 3.55 & 1.20 & 1.46 & 1.88 \\
AnyEdit~\citep{yu2025anyedit} & 3.18 & 2.95 & 1.88 & 2.47 & 2.23 & 2.24 & 2.85 & 1.56 & 2.65 & 2.45 \\
UltraEdit~\citep{zhao2024ultraedit} & 3.44 & 2.81 & 2.13 & 2.96 & 1.45 & 2.83 & 3.76 & 1.91 & 2.98 & 2.70 \\
OmniGen~\citep{xiao2025omnigen} & 3.47 & 3.04 & 1.71 & 2.94 & 2.43 & 3.21 & 4.19 & 2.24 & 3.38 & 2.96 \\
ICEdit~\citep{zhang2025context} & 3.58 & 3.39 & 1.73 & 3.15 & 2.93 & 3.08 & 3.84 & 2.04 & 3.68 & 3.05 \\
Step1X-Edit~\citep{liu2025step1x} & 3.88 & 3.14 & 1.76 & 3.40 & 2.41 & 3.16 & 4.63 & 2.64 & 2.52 & 3.06 \\
BAGEL~\citep{deng2025emerging} & 3.56 & 3.31 & 1.70 & 3.3 & 2.62 & 3.24 & 4.49 & 2.38 & 4.17 & 3.20 \\
UniWorld-V1~\citep{lin2025uniworld} & 3.82 & 3.64 & 2.27 & 3.47 & 3.24 & 2.99 & 4.21 & 2.96 & 2.74 & 3.26 \\
OmniGen2~\citep{wu2025omnigen2} & 3.57 & 3.06 & 1.77 & 3.74 & 3.20 & 3.57 & 4.81 & 2.52 & 4.68 & 3.44 \\
FLUX.1 Kontext [Pro]~\citep{labs2025flux1kontextflowmatching}  & 4.25 & 4.15 & 2.35 & 4.56 & 3.57 & 4.26 & 4.57 & 3.68 & 4.63 & 4.00 \\
GPT-Image-1 [High]~\citep{openai_image_api} & \textbf{4.61} & 4.33 & 2.90 & 4.35 & 3.66 & \textbf{4.57} & \textbf{4.93} & \textbf{3.96} & \textbf{4.89} & 4.20 \\
Qwen-Image-Edit~\citep{wu2025qwen} & 4.38 & 4.16 & 3.43 & 4.66 & 4.14 & 4.38 & 4.81 & 3.82 & 4.69 & 4.27 \\
\midrule
FLUX.1 Kontext [Dev]~\citep{labs2025flux1kontextflowmatching} & 4.12 & 3.80 & 2.04 & 4.22 & 3.09 & 3.97 & 4.51 &3.35 & 4.25 & 3.71 \\
UniWorld-FLUX.1-Kontext & 4.19 & 4.20 & 2.43 & 4.32 & 3.91 & 4.08 & 4.68 & 3.72 & 4.63 & 4.02 \\
\rowcolor{green!5}
\textit{vs. Baseline} & \textcolor{deepgreen}{\textbf{+0.07}} & 
\textcolor{deepgreen}{\textbf{+0.40}} & 
\textcolor{deepgreen}{\textbf{+0.39}} & 
\textcolor{deepgreen}{\textbf{+0.10}} & 
\textcolor{deepgreen}{\textbf{+0.82}} & 
\textcolor{deepgreen}{\textbf{+0.11}} & 
\textcolor{deepgreen}{\textbf{+0.17}} & 
\textcolor{deepgreen}{\textbf{+0.37}} & 
\textcolor{deepgreen}{\textbf{+0.38}} & 
\textcolor{deepgreen}{\textbf{+0.31}} \\
\midrule
Qwen-Image Edit [2509]~\citep{wu2025qwen} & 4.32 & 4.36     & 4.04 & 4.64 & 4.52 & 4.37 & 4.84 & 3.39 & 4.71 & 4.35 \\
UniWorld-Qwen-Image-Edit & 4.34 & 4.40 & \textbf{4.37} & 4.62 & 4.65 & 4.31 & 4.89 & 3.90 & 4.80 & 4.48 \\
\rowcolor{green!5}
\textit{vs. Baseline} & \textcolor{deepgreen}{\textbf{+0.02}} & 
\textcolor{deepgreen}{\textbf{+0.04}} & 
\textcolor{deepgreen}{\textbf{+0.33}} & 
\textcolor{gray}{\textbf{-0.02}} & 
\textcolor{deepgreen}{\textbf{+0.13}} & 
\textcolor{gray}{\textbf{-0.06}} & 
\textcolor{deepgreen}{\textbf{+0.05}} & 
\textcolor{deepgreen}{\textbf{+0.51}} & 
\textcolor{deepgreen}{\textbf{0.09}} & 
\textcolor{deepgreen}{\textbf{+0.13}} \\
\midrule
UniWorld-V2 & 4.29 & \textbf{4.44 }& 4.32 & \textbf{4.69} & \textbf{4.72} & 4.41 & 4.91 & 3.83 & 4.83  & \textbf{4.49} \\
\bottomrule[1.5pt]
\end{tabular}}
\label{tab_ImgEdit} 
\end{table*}

\subsection{Experimental Setup}

To evaluate the effectiveness of our approach, we conduct experiments from two perspectives: 1) The alignment between different MLLM scoring methods and human judgments, and 2) The performance improvement of the editing model after post-training with our method.

\textbf{Training}\quad We use FLUX.1-Kontext [Dev]~\citep{labs2025flux1kontextflowmatching}, Qwen-Image-Edit [2509]~\citep{wu2025qwen} and UniWorld-V2 as our base models. For training, we allocate 3 nodes to FLUX.1-Kontext [Dev], 6 nodes to Qwen-Image-Edit [2509] and 9 nodes to UniWorld-V2, with each node containing 8 A100 GPUs. We perform MLLM scoring on a single node using vLLM~\citep{kwon2023efficient}. To optimize GPU memory utilization, we employ Fully Sharded Data Parallelism (FSDP) for the text encoder and gradient checkpointing when training Qwen-Image-Edit [2509] and UniWorld-V2.

\textbf{Evaluation}\quad For quantitative evaluation, we employ two comprehensive benchmarks: ImgEdit~\citep{ye2025imgedit}, which unifies multiple specialized tasks into a common framework for comprehensive model comparison, and GEdit-Bench~\citep{liu2025step1x}, which assesses general-purpose image editing through diverse natural language instructions.

\subsection{Main Results}
\begin{figure}[ht]
    \centering
        \begin{minipage}[c]{0.55\textwidth}
        \centering
        \captionsetup{type=table} 
        \caption{Quantitative comparison on GEdit-Bench~\citep{liu2025step1x}. \textbf{Bold} indicates the best performance.}
        \label{tab_GEdit}
        \scalebox{0.75}{
        \renewcommand{\arraystretch}{1.0}
        \setlength{\tabcolsep}{10pt}
        \begin{tabular}{l|ccc}
        \toprule[1.5pt]
        \textbf{Model} & \multicolumn{3}{c}{\textbf{GEdit-Bench-EN} $\uparrow$} \\
        \cmidrule(lr){2-4}
         & \textbf{G\_SC} & \textbf{G\_PQ} & \textbf{G\_O} \\
        \midrule
        Instruct-Pix2Pix & 3.58 & 5.49 & 3.68 \\
        AnyEdit & 3.18 & 5.82 & 3.21 \\
        MagicBrush & 4.68 & 5.66 & 4.52 \\
        UniWorld-V1 & 4.93 & 7.43 & 4.85 \\
        OmniGen & 5.96 & 5.89 & 5.06 \\
        FLUX.1-Kontext [Dev] &  6.52 & 7.38 & 6.00 \\
        OmniGen2 & 7.16 & 6.77 & 6.41 \\
        Gemini 2.0 & 6.73 & 6.61 & 6.32 \\
        BAGEL & 7.36 & 6.83 & 6.52 \\
        FLUX.1-Kontext [Pro] & 7.02 & 7.60 & 6.56 \\
        Step1X-Edit & 7.66 & 7.35 & 6.97 \\
        UniPic2 & - & - & 7.10 \\
        GPT-Image-1 [High] & 7.85 & 7.62 & 7.53 \\
        Qwen-Image-Edit & 8.00 & 7.86 & 7.56 \\
        \midrule
        FLUX.1-Kontext [Dev] &  6.52 & 7.38 & 6.00 \\
        UniWorld-FLUX.1-Kontext & 7.28 & 7.49 & 6.74 \\
        \rowcolor{green!5}
        \textit{vs. Baseline} & \textcolor{deepgreen}{\textbf{+0.76}} & \textcolor{deepgreen}{\textbf{+0.11}} & \textcolor{deepgreen}{\textbf{+0.74}} \\
        \midrule
        Qwen-Image-Edit [2509] & 8.15 & 7.86 & 7.54 \\
        UniWorld-Qwen-Image-Edit & 8.36 & 7.87 & 7.76 \\
        \rowcolor{green!5}
        \textit{vs. Baseline} & \textcolor{deepgreen}{\textbf{+0.21}} & \textcolor{deepgreen}{\textbf{+0.01}} & \textcolor{deepgreen}{\textbf{+0.22}} \\
        \midrule
        UniWorld-V2 & \textbf{8.39} & \textbf{8.02} & \textbf{7.83} \\
        \bottomrule[1.5pt]
        \end{tabular}}
    \end{minipage}
    \hfill
    \begin{minipage}[c]{0.43\textwidth}
        \centering
        \includegraphics[width=0.95\linewidth]{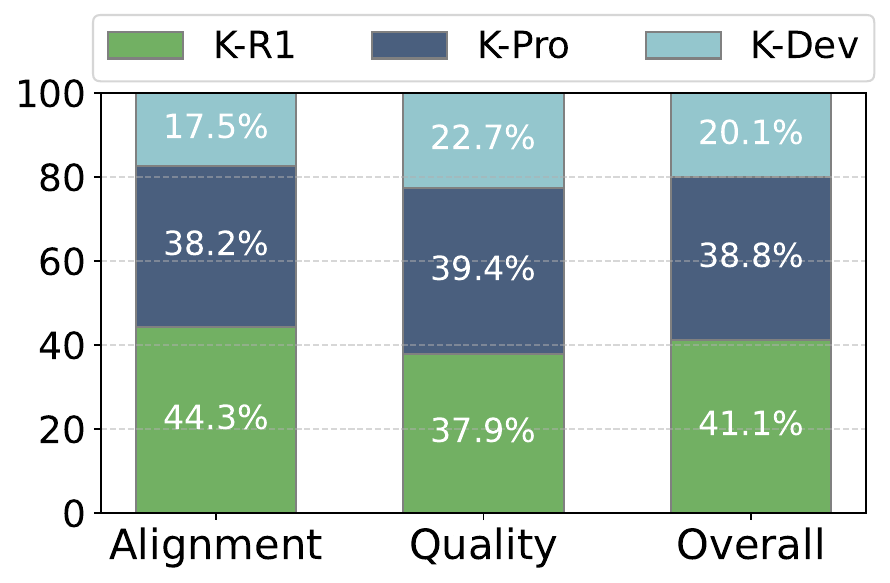}
        \vspace{-1em}
        \hspace{0.2em}
        \includegraphics[width=\linewidth]{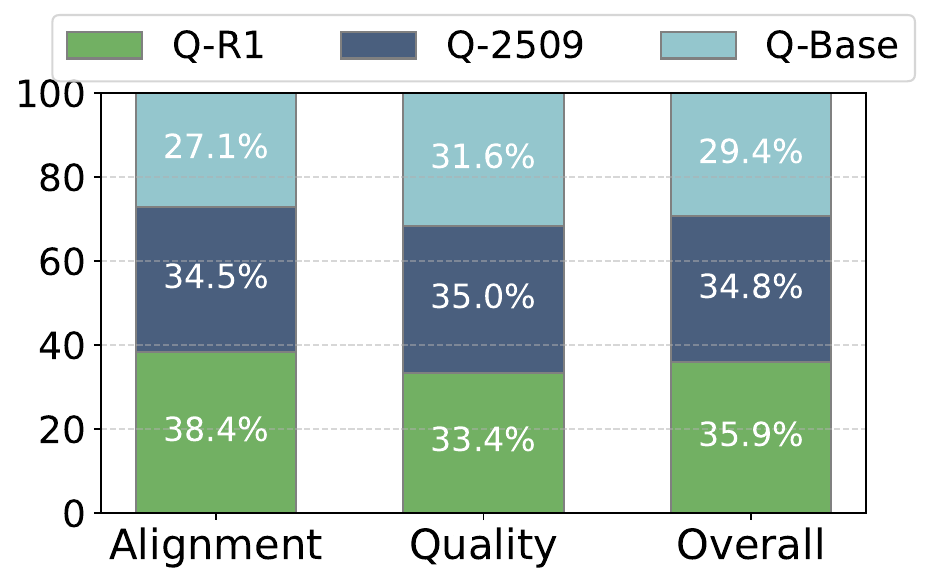}
        \caption{User study for Kontext-based (up) and Qwen-based (bottom) model.}
        \label{fig:combined_results}
    \end{minipage}

\end{figure}

We evaluate these models on the ImgEdit and GEdit-Bench benchmarks to assess their editing capabilities and generalization. The quantitative results are presented in~\autoref{tab_ImgEdit} and~\autoref{tab_GEdit}, respectively, and a qualitative comparison is shown in~\autoref{fig_case}.

\textbf{Our method unlocks the model's potential and significantly improves its performance.}\quad As shown on the ImgEdit benchmark in \autoref{tab_ImgEdit}, our method substantially enhances the performance of all base models. For FLUX.1-Kontext [Dev], the overall score improves significantly from 3.71 to 4.02, outperforming the stronger Pro version (4.00). Similarly, when applied to Qwen-Image-Edit [2509], our method boosts its score from 4.35 to an impressive 4.48, achieving state-of-the-art performance among open-source models and surpassing top-tier closed-source models like GPT-Image-1. Beyond the gains in the overall score, a dramatic performance surge is observed in the `Adjust', `Extract', and `Remove` dimensions for UniWorld-FLUX.1-Kontext and in `Extract' and `Hybrid' dimensions for UniWorld-Qwen-Image-Edit. Moreover, UniWorld-V2 achieves the best performance. This phenomenon indicates that our method can unlock and significantly improve the previously underdeveloped potential within the base models.

\textbf{Our method exhibits robust generalization capabilities on the out-of-domain dataset.}\quad On the out-of-domain GEdit-Bench~(\autoref{tab_GEdit}), Edit-R1 demonstrates strong generalization for three models. It enhances the FLUX.1-Kontext [Dev] model's overall score from 6.00 to 6.74, yielding a performance that surpasses the Pro version (6.56). For the Qwen-Image model, the score is increased from 7.54 to 7.76. Meanwhile, UniWorld-V2 establishes a new state-of-the-art on this benchmark by outperforming all listed models, including Qwen-Image-Edit (7.56) and GPT-Image-1 (7.53). This result confirms that our method effectively preserves and enhances core editing capabilities on unseen data distributions, showcasing strong generalization.

\textbf{Our method proves its effectiveness in human preference evaluations.} \quad For a comprehensive evaluation, we conducted a human preference study for both the FLUX.1 and Qwen series, where participants compared our finetuned model with its base models and the more powerful version. They were asked to select the best result across two dimensions: instruction alignment and image quality. As detailed in~\autoref{fig:combined_results}, users prefer the UniWorld-FLUX.1-Kontext over FLUX.1-Kontext [Dev] across all criteria. Furthermore, it demonstrates stronger editing capabilities when compared to the more powerful official versions, FLUX.1-Kontext [Pro]. Overall, the UniWorld-FLUX.1-Kontext gains more likes  due to its superior instruction-following ability, even though the official models are slightly superior in image quality. This confirms that our method effectively steers the model to generate outputs that better align with human preferences.

\begin{figure}[ht]
    \centering
    \begin{minipage}{0.46\textwidth}
        \centering
        \includegraphics[width=0.87\linewidth]{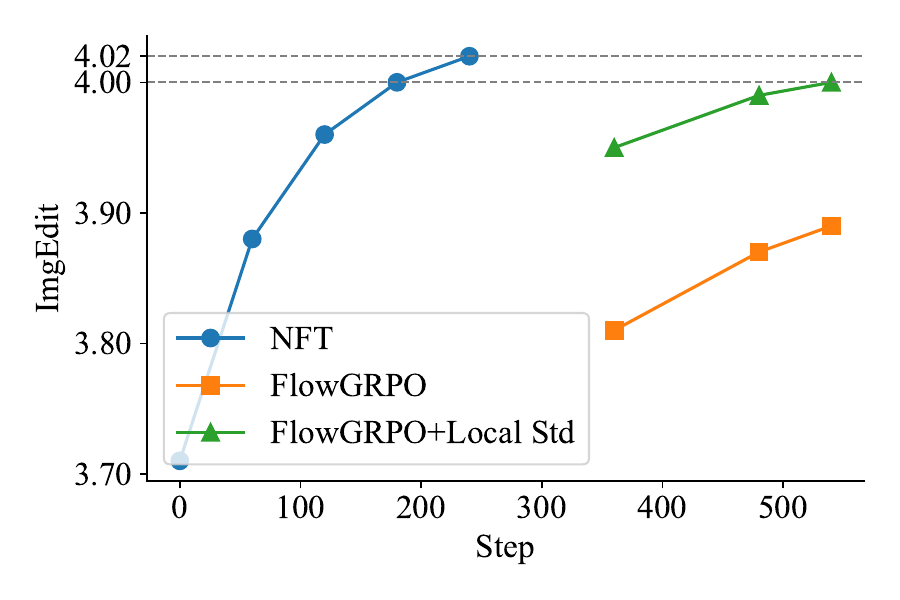} 
        \vspace{-1.5em}
        \caption{Results of different policy optimization methods on FLUX.1-Kontext [Dev].}
        \label{fig:ablation_methods} 
    \end{minipage}
    \hfill
    \begin{minipage}{0.52\textwidth}
        \centering
        \captionsetup{type=table}
        \captionof{table}{Ablation study of our core components using Qwen-Image-Edit [2509] on GEdit-Bench.}
        \label{ablation}
        \vspace{0.5em}
        \scalebox{0.95}{
            \begin{tabular}{l|c}
                \toprule[1.5pt]
                \textbf{Model} & \textbf{GEdit-Bench $\uparrow$} \\
                \midrule
                \textbf{Qwen-Image-Edit [2509]} & 7.54 \\
                \midrule
                + NFT (7B) & 7.66 \\
                + 32B & 7.74 \\
                + Group Filtering & 7.76 \\
                \bottomrule[1.5pt]
            \end{tabular}
        }
        \label{fig:ablation_study}
    \end{minipage}
    \hfill
\end{figure}

\subsection{Ablation Study}
We conducted ablation studies to validate our core components. As presented in~\autoref{fig:ablation_methods}, we employ DiffusionNFT as the policy optimization method on FLUX.1 Kontext [Dev]. It achieves superior performance on the ImgEdit benchmark over baselines, including Flow-GRPO and its variant using local std. Moreover, as shown in~\autoref{fig:ablation_study}, applying DiffusionNFT to the Qwen-Image-Edit [2509] baseline model significantly raises its score on GEdit-Bench from 7.54 to 7.72. Introducing the group filtering mechanism further increases the score to 7.76. 

\subsection{Analysis}

\begin{figure}[ht]
    \centering
    \begin{minipage}{0.45\textwidth}
        \centering
        \includegraphics[width=1.0\linewidth]{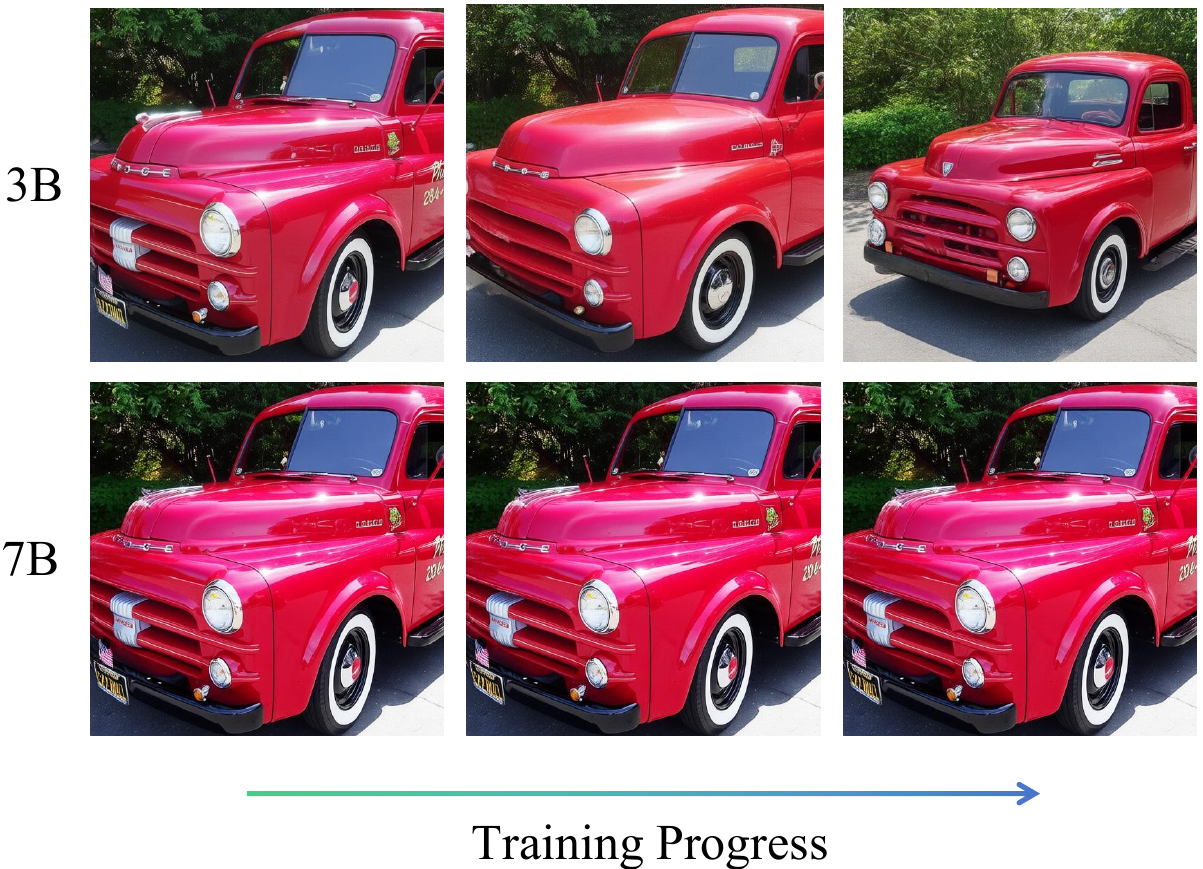} 
        \label{fig:reward_hacking} 
    \end{minipage}
    \hfill
    \begin{minipage}{0.53\textwidth}
      \centering
        \includegraphics[width=\linewidth]{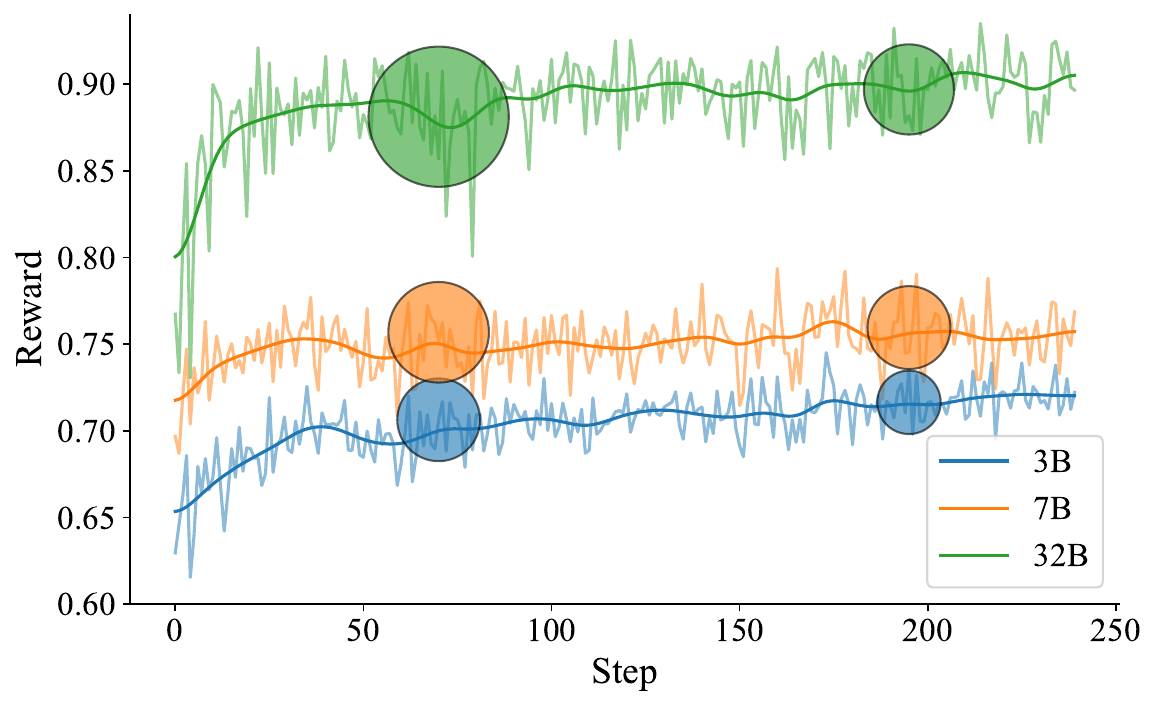} 
        \label{fig:reward_dynamic}
    \end{minipage}
     \caption{Reward hacking phenomenon observed in the 3B reward model (Left) and Training reward dynamics across varying reward model scales (Right).}
    \label{fig:hack_all}
\end{figure}

\begin{figure}[ht]
    \centering
    \begin{minipage}{0.48\textwidth}
      \centering
        \captionsetup{type=table} 
  \caption{Pairwise accuracy of different reward methods against human preferences}
        \label{tab_pairwise_accuracy}
        \setlength{\tabcolsep}{1pt}
        \renewcommand{\arraystretch}{1.3}
        \scalebox{1}{
        \begin{tabular}{lc}
        \toprule[1.5pt]
        \textbf{Reward Method} & \textbf{Accuracy (\%)} \\
        \midrule
        Score Sampling (S\_S)           & 60.82 \\
        Yes/No Logit (Y/N\_L)         & 67.01 \\
        Score Sampling + CoT   & 62.37 \\
        Score Logit + CoT   & 63.40 \\
        \makecell[l]{UnifiedReward (UR)\\\citep{wang2025unified}}            & 65.46 \\
        \midrule
        \textbf{Score Logit (Ours)} & \textbf{74.74} \\
        \bottomrule[1.5pt]
        \end{tabular}
        }
    \end{minipage}
    \hfill
        \begin{minipage}{0.44\textwidth}
        \centering
        \includegraphics[width=\linewidth]{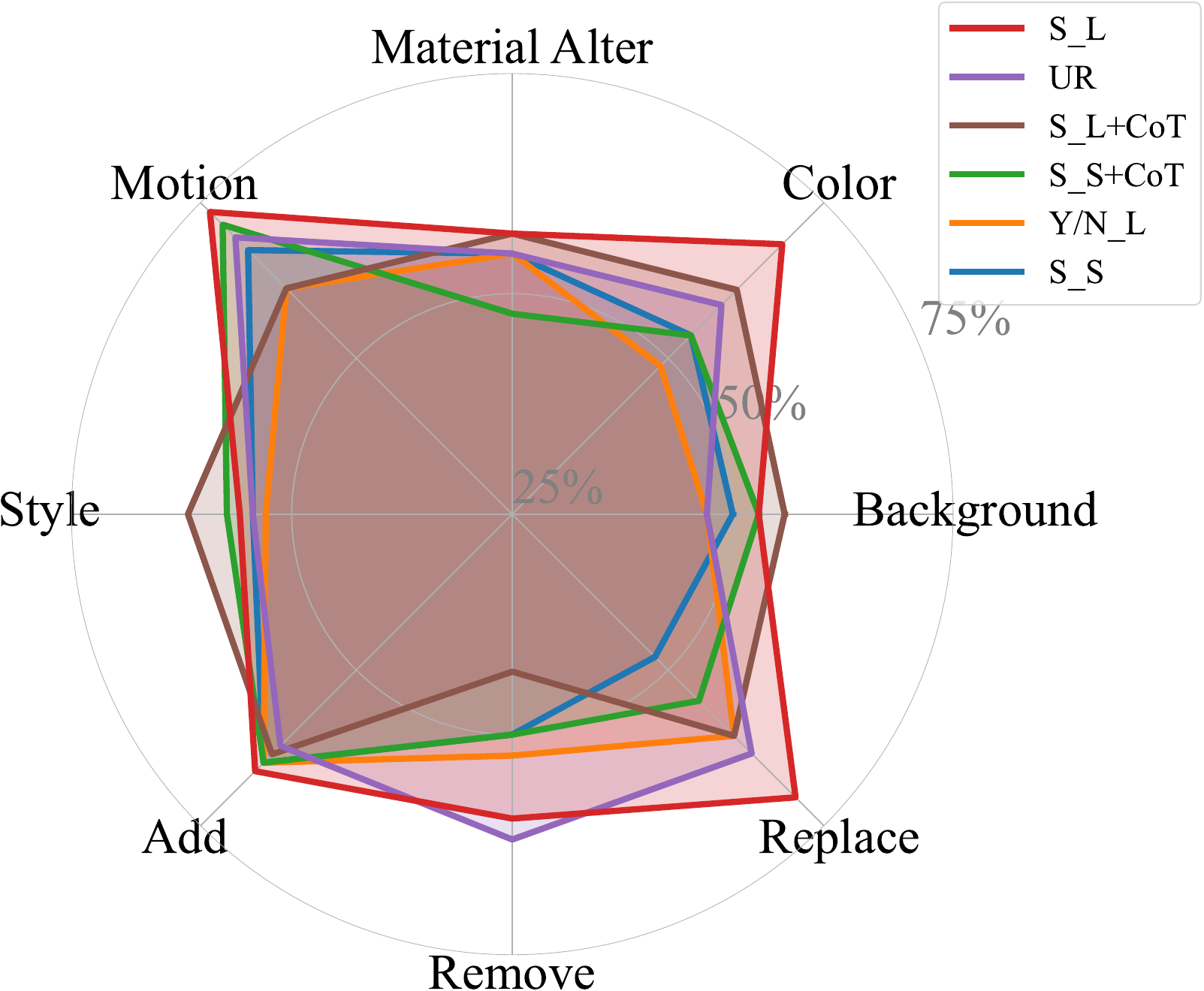} 
        \caption{Performance comparison across different editing tasks.} 
        \label{fig_radar_chart_cate} 
    \end{minipage}
\end{figure}

\textbf{Human Alignment.}\quad 
To validate our choice of reward mechanism, we evaluate the alignment between different scoring methods and human judgments. The results indicate that our adopted logit-based method achieves the highest correlation with human preferences among all evaluated reward mechanisms. As detailed in~\autoref{tab_pairwise_accuracy}, this method achieves an overall pairwise accuracy of 74.74\%, significantly outperforming other methods. Furthermore, the results in~\autoref{fig_radar_chart_cate} demonstrate that superior alignment is consistent across diverse editing tasks. We provide a detailed analysis of these findings in~\autoref{app_align}, including score distributions and the detailed experimental protocol.

\textbf{Reward Model Scaling.}\quad
To evaluate the effect of reward model scaling on policy model performance, we fine-tuned Qwen-Image-Edit for the same number of steps using reward models of varying parameter scales for a fair comparison. As shown in \autoref{ablation}, with an increase in the scale of the reward model, the overall score of the policy model improves, demonstrating that scaling the reward model contributes to continuous performance improvement. 

\textbf{Reward Hacking and Reward Variance.}\quad 
As illustrated in~\autoref{fig:hack_all} (Left), the policy model fine-tuned on 3B model exhibits significant reward hacking and its edit results deviate from the source image. In contrast, the model fine-tuned on larger 7B model alleviates this issue.
For further investigation, we analyze the training reward curves and attribute the phenomenon to the variance of reward scores. As depicted in~\autoref{fig:hack_all} (Right), we visualize the smoothed reward trajectory (solid line), raw reward fluctuations (shaded line), and the reward variance (bubble size) as an indicator of exploration intensity.

Our observations are as follows:
i) \textit{Reward Hacking in Smaller Models}: Smaller reward model, such as 3B and 7B, exhibit reward hacking, as their reward variance rapidly collapses early in training, indicating a premature halt to effective exploration.
ii) \textit{Sustained Exploration in Larger Models}: In contrast, the 32B model maintains high reward variance throughout training, demonstrating sustained exploration capability which enables the discovery of superior solutions even in later stages.
This phenomenon is also analyzed in another study~\citep{wu2025rewarddance}.
These dynamics suggest that scaling reward models can effectively mitigate reward hacking and maintain robust exploration.
\section{Related Work}
\subsection{Image Editing}
The advent of diffusion models has marked a pivotal moment in Text-to-Image (T2I) generation~\citep{song2020score, rombach2022high, lipman2022flow, liu2025cot, yan2025gpt}. Image editing presents a more constrained challenge: altering specific attributes while preserving unedited regions. Early methods like SDEdit~\citep{meng2021sdedit} offered global stylistic control but lacked spatial precision. To address this, inversion-based techniques~\citep{hertz2022prompt} were developed to reconstruct an image from its latent representation, with methods like Null-text Inversion~\citep{mokady2023null} optimizing the process, though they can be computationally intensive and may introduce artifacts~\citep{huang2025diffusion}. Other methods focus on adding explicit conditional control, such as ControlNet~\citep{zhang2023adding} for spatial guidance and IP-Adapter~\citep{ye2023ip} for image-based prompting. Concurrently, training-based fine-tuning approaches~\citep{brooks2023instructpix2pix} adapted models for editing but faced generalization issues. Building on these foundations, a new wave of powerful instruction-tuned models such as ICEdit~\citep{zhang2025context} and Step1X-Edit~\citep{liu2025step1x} has emerged. These are increasingly integrated with general-purpose MLLMs, leading to highly capable systems like BAGEL~\citep{deng2025emerging}, Qwen-Image~\citep{wu2025qwen}, and GPT-Image-1~\citep{openai_image_api}, while foundational architectures also evolve with frameworks like flow matching in FLUX Kontext~\citep{labs2025flux1kontextflowmatching}.

\subsection{Reinforcement Learning in Generative Models}
Reinforcement Learning From Human Feedback (RLHF)~\citep{ouyang2022training} has become the dominant paradigm for aligning LLMs to be more helpful~\citep{hu2022lora, shao2024deepseekmath}, honest~\citep{gao2024honestllm}, and harmless~\citep{yang2025asft}. Inspired by this success, researchers have adapted the RL framework to T2I models~\citep{black2023training}, typically by training a reward model (RM) on general human preferences~\citep{xu2024visionreward} or specific prompt-image alignment scores~\citep{xu2023imagereward}. However, this ``train-an-RM-then-RL'' pipeline is suboptimal for image editing, as RMs are difficult to craft~\citep{yang2025rlir, miao2024training}. Alternatives like Direct Preference Optimization (DPO)~\citep{rafailov2023direct} remove the explicit RM but rely on static, offline data, which provides a less dynamic signal than required for iterative visual refinement. Similarly, using powerful MLLMs to provide discrete scores~\citep{gong2025onereward, niu2025wise} yields a coarse signal that fails to capture the continuous nature of visual quality. More advanced algorithms like GRPO~\citep{shao2024deepseekmath} have also shown promise in aligning both diffusion and flow matching models, as seen in FlowGRPO~\citep{liu2025flow},  which was first utilized in Skywork UniPic 2.0~\citep{wei2025skywork} for image editing,  and DanceGRPO~\citep{xue2025dancegrpo}, but can still be exploited via reward hacking~\citep{wang2025pref}. Differing from these policy gradient approaches, DiffusionNFT~\citep{zheng2025diffusionnft} presents an online RL paradigm that directly optimizes the model on the forward process via a negative-aware fine-tuning objective.

\subsection{MLLMs as Evaluators and Reward Models}
The emergence of powerful MLLMs has established the ``MLLM-as-a-Judge'' paradigm~\citep{chen2024mllm}, demonstrating a high correlation with human judgments~\citep{zhang2025upme}. This has motivated a shift from passive evaluation to using MLLMs as an active reward to optimize generative models~\citep{gong2025onereward, xu2023imagereward,jin2025srum,niu2025wise}. However, converting MLLM evaluations into an effective reward signal for image editing presents challenges. A straightforward approach of using discrete scores~\citep{gong2025onereward} provides a sparse and coarse signal, ill-suited for the continuous and subtle nature of visual improvements. To obtain finer-grained reward signals, logit-based scoring methods generate rewards by computing the expected value of the token distribution from model outputs~\citep{wu2024q, zhang2024q, li2025generalist}. Another strategy involves learning from large, static pairwise preference datasets~\citep{xu2023imagereward}, which are often used in alignment algorithms like DPO adapted for T2I models~\citep{black2023training}. This offline method decouples the feedback from the live generation process, potentially failing to cover the vast distribution of possible edits and lacking the real-time, iterative guidance needed for optimization~\citep{zhang2024large}. 
The goal is to develop a feedback mechanism that can effectively guide editing models that are already leveraging MLLMs for planning and reasoning~\citep{liu2025cot, yang2024mastering}.
Another feedback mechanism to guide editing models involves leveraging MLLMs for planning and reasoning~\citep{liu2025cot, yang2024mastering}, offering more direct and interpretable guidance than a simple feedback score.

\section{Conclusion}
In this paper, we introduce Edit-R1, a novel post-training framework designed to overcome generalization limitations in instruction-based image editing models. Our core innovation is using an MLLM as a training-free reward model, which provides fine-grained, continuous feedback directly from its output logits, combined with the efficient DiffusionNFT, a likelihood-free policy optimization method consistent with the flow matching forward process. Extensive experiments demonstrate that our framework achieves state-of-the-art performance on the ImgEdit and GEdit-Bench by significantly boosting various base models, including UniWorld-V2, FLUX.1-Kontext, and Qwen-Image-Edit. Our analysis confirms that the MLLM-derived reward signal exhibits a high correlation with human preferences, effectively guiding the model towards higher-quality outputs while mitigating reward hacking.

\section{Contributors}

\textbf{Core Contributors}: Zongjian Li, Zheyuan Liu, Qihui Zhang, Bin Lin

\textbf{Contributors}: Feize Wu, Shenghai Yuan, Zhiyuan Yan, Yang Ye, Wangbo Yu, Yuwei Niu, Shaodong Wang, Xinhua Cheng

\textbf{Corresponding Author}: Li Yuan

{
    \tt
    \{zongjianli25@stu., yuanli-ece@\}pku.edu.cn
}

\bibliography{iclr2026_conference}
\bibliographystyle{iclr2026_conference}

\newpage
\appendix

\section{Training Settings}
\label{Training_settings}

We present key hyperparameters of the training in \autoref{tb:train_hyper}.

\begin{table}[ht]
\centering
\renewcommand{\arraystretch}{1.0}
\setlength{\tabcolsep}{40pt}
\begin{tabular}{cc}
\toprule[1.2pt]
\textbf{Parameter} & \textbf{Setting} \\
\midrule
\multicolumn{2}{c}{\textit{Basic}} \\
\midrule
Learning Rate & 3e-4 \\
$\beta_1$ & 0.9 \\
$\beta_2$ & 0.999 \\
Batch Size & 3 \\
EMA Decay & 0.9 \\
\midrule
\multicolumn{2}{c}{\textit{Sampling}} \\
\midrule
Sampling Inference Steps & 6 \\
Resolution & 512 $\times$ 512 \\
The Number of Images Per Prompt & 12 \\
The Number of Groups & 24 \\
\midrule
\multicolumn{2}{c}{\textit{DiffusionNFT}} \\
\midrule
KL Loss Weight & 0.0001 \\ 
Guidance Strength ($\frac{1}{\beta}$) & 1.0 \\
\midrule
\multicolumn{2}{c}{\textit{Group Filtering}} \\
\midrule
$\tau_\mu$ & 0.9 \\ 
$\tau_\sigma$ & 0.05 \\
\bottomrule[1.2pt]
\end{tabular}
\caption{Key hyperparameters.}
\label{tb:train_hyper}
\end{table}

\section{Detailed Human Alignment}
\label{app_align}

We conducted a comprehensive human alignment study to rigorously evaluate the extent to which our proposed reward mechanism aligns with human judgment. We analyze this alignment from two perspectives: the accuracy of pairwise comparisons and the similarity of score distributions. For this part, we collected 800 edited images generated from 200 unique image-instruction pairs. These results were then annotated by three human evaluators across two distinct dimensions.

\subsection{Reward Method Definitions}
\label{reward_def}
To comprehensively evaluate the human alignment of various reward mechanisms, we compared our proposed ``Score Logit'' method against several baselines. The details of each method are as follows:

\textbf{Score Logit (Ours):} This is our primary reward mechanism. We prompt the MLLM to evaluate the edit on a scale of 0 to 5. We then extract the logits corresponding to the score tokens (i.e., ``0'', ``1'', ``2'', ``3'', ``4'', ``5''), apply a softmax function to obtain a probability distribution, and compute the expected value (weighted average) as the raw reward score. This score is subsequently normalized to the range of [0, 1]. This process is detailed in~\autoref{mllmscore}.

\textbf{Score Sampling (S\_D):} In this approach, the MLLM is prompted to directly output a single numerical score from 1 to 5. This explicit score is then parsed and normalized to a value between 0 and 1 to serve as the reward.

\textbf{Yes/No Logit (Y/N\_C):} This method reframes the evaluation as a binary classification task. We prompt the MLLM to determine if the edit was successful (``Yes'' or ``No''). We then extract the logits for the ``Yes'' and ``No'' tokens, apply a softmax function, and use the resulting probability of ``Yes'' as the final reward signal.

\textbf{Score Sampling + CoT:} This is a variant of the discrete scoring method that incorporates Chain-of-Thought (CoT) reasoning. The MLLM is first instructed to generate a brief analysis of the image edit before outputting the final 0-5 score. The reward is derived from this final score, normalized to [0, 1].

\textbf{Score Logit + CoT:} Similar to the above, this method adds a CoT step to our continuous scoring approach. The MLLM first provides its reasoning, and then we extract the logits for the score tokens that follow the reasoning to calculate the weighted average reward.

\textbf{Unified Reward (UR):} As a strong baseline, we utilize the pre-trained reward model from the work of~\citet {wang2025unified}. We use the direct output of this model as the reward signal, without any modification or re-training.

\subsection{Alignment in Pairwise Preference}

First, we assess the alignment in terms of pairwise preference. For each input condition (original image and instruction), human annotators were asked to compare pairs of edited images and determine which one was better, or if they were of equivalent quality. This human judgment serves as the ground truth. We then measure the alignment of different reward models by calculating their pairwise accuracy—the percentage of pairs where the model's preference (i.e., which image receives a higher reward) matches the human preference. A higher accuracy indicates that the reward model's perception of relative quality is more similar to that of humans.

\begin{figure}[ht]
    \centering
    \begin{minipage}{0.48\textwidth}
        \centering
        \includegraphics[width=\linewidth]{figure/model_performance_radar_chart.pdf} 
        \caption{Performance comparison across different editing tasks.} 
        \label{fig_radar_chart_cate2} 
    \end{minipage}
    \hfill
    \begin{minipage}{0.48\textwidth}
        \centering
        \includegraphics[width=\linewidth]{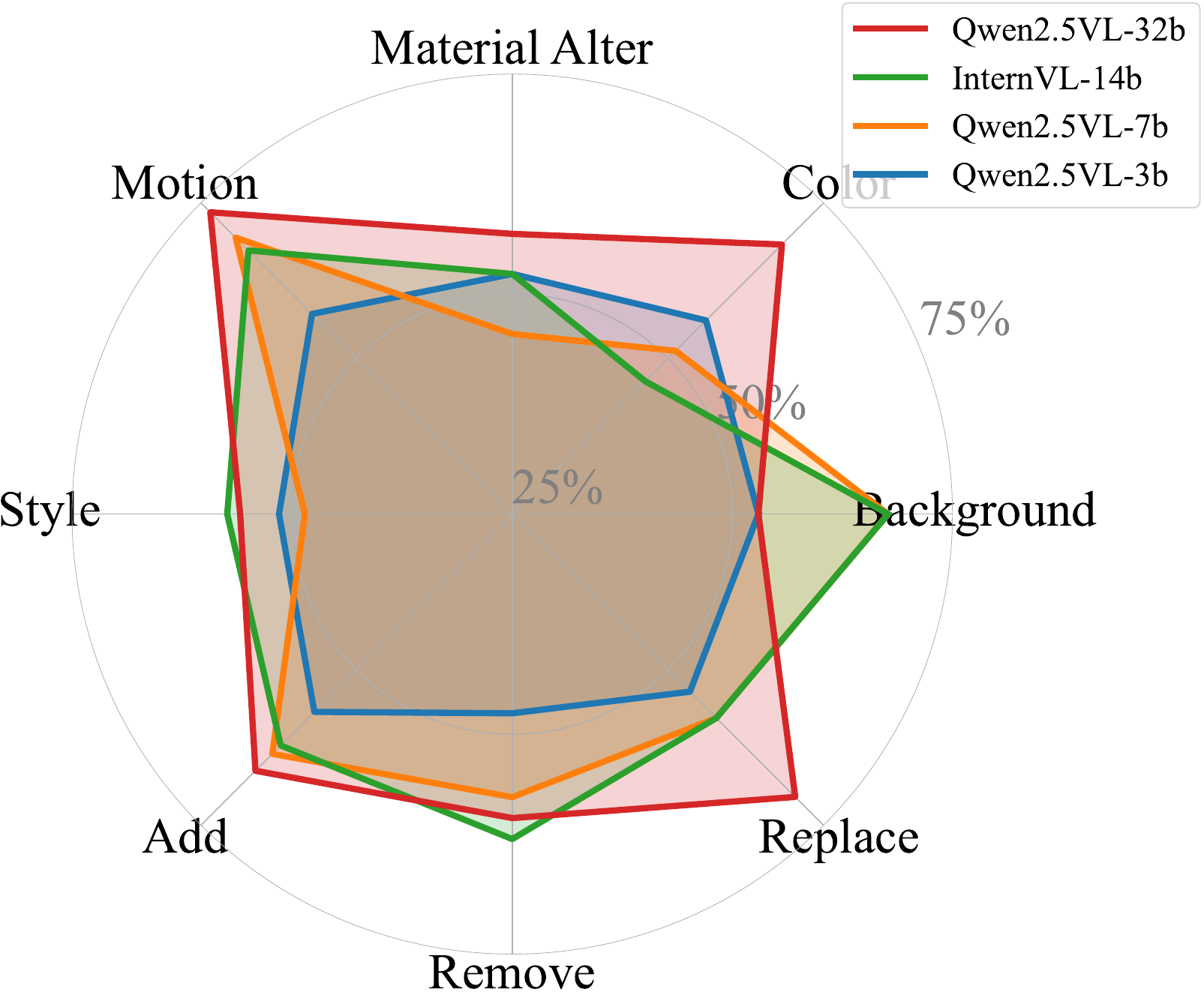} 
        \caption{Performance comparison across different models.} 
        \label{fig_radar_chart_model} 
    \end{minipage}
\end{figure}

The results are presented in~\autoref{fig_radar_chart_cate2}. Our proposed method, ``Score Logit'', which utilizes the expected value of score logits, achieves a pairwise accuracy of 74.74\%. This result significantly surpasses all other baseline methods, including binary classification-based rewards and those using discrete scores. This demonstrates that our continuous reward signal is more effective at capturing the nuanced differences in edit quality that align with human perception.

We observe that incorporating Chain-of-Thought (CoT) unexpectedly degrades the performance of continuous scoring methods. We attribute it to a reasoning-induced bias, akin to the exposure bias in sequence generation~\citep{pozzi2025mitigating}. Recent studies suggest that when a model generates a lengthy reasoning chain, it can become overly persuaded by its own internally generated narrative, even if that narrative is flawed or drifts away from the initial evidence~\citep{wu2023self}. In the context of image editing evaluation, the CoT process may cause the MLLM to focus more on its textual justification rather than the subtle but critical visual details of the images. Consequently, its final judgment becomes less grounded. We enforce a direct evaluation with a restricted output format, which effectively mitigates this bias and ensures the assessment remains tightly coupled with the visual input.

\subsection{Alignment in Score Distribution}

\begin{figure}[ht]
    \centering
    \includegraphics[width=0.6\linewidth]{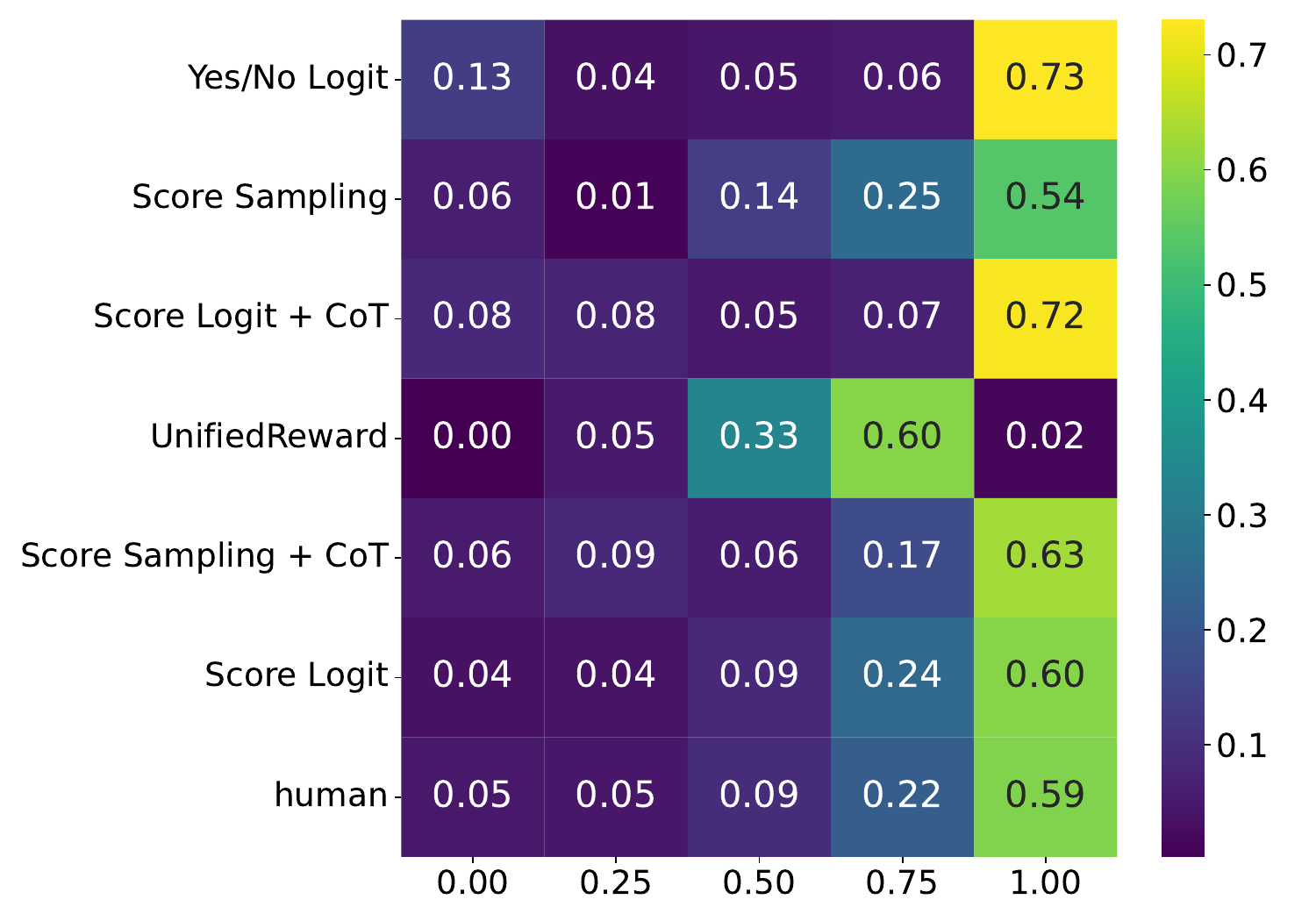}
    \caption{Score distribution across different reward methods}
    \label{fig_heatmap}
\end{figure}

We analyze the alignment of score distributions. In this task, annotators assigned a quality label (``Good'', ``Bad'', or ``Indistinguishable'') and an absolute quality score from 1 to 5 to each edited image. 
We evaluate alignment from two aspects. First, we compare the overall distribution of scores generated by each reward model against the distribution of scores given by human annotators. As shown in~\autoref{fig_heatmap}, a high degree of similarity in distribution suggests that the reward model shares a similar tendency and preference scale with humans. 

The results clearly indicate the superiority of ``Score Logit''. The figures show that its score distribution most closely mirrors that of human evaluators. Furthermore, it exhibits strong consistency with human judgment, assigning significantly higher scores to ``Good'' edits and lower scores to ``Bad'' ones compared to the other methods. Both analyses confirm that ``Score Logit'' provides a reward signal that is more accurate in relative comparisons and better calibrated to the absolute scale of human quality perception, leading to a higher degree of alignment.

\subsection{Annotation Details}
\label{App_anno}

To construct a robust ground truth for evaluating the human alignment of our reward mechanism, we undertook a detailed manual annotation process. We collected 800 edited images generated from 200 unique image-instruction pairs for this study. The annotation was carried out by a team of five human evaluators, all of whom are proficient in English and possess a strong understanding of the image editing domain.

To ensure the reliability and consistency of our annotations, we implemented a rigorous quality control protocol. Each annotation task was initially assigned to three distinct evaluators to ensure multiple perspectives. A label was accepted if a consensus was reached, requiring at least two of the three annotators to provide the same judgment. In cases where no initial consensus was reached, the task was reassigned to two of the remaining evaluators who had not previously assessed the sample. A final label was considered valid only if an overall agreement rate of 60\% or higher was achieved across all assigned evaluators. Samples that failed to meet this threshold were discarded from our analysis to maintain the high quality and integrity of the ground-truth data.

The evaluators performed two primary annotation tasks corresponding to the dimensions of our human alignment analysis. For pairwise preference comparison, annotators were presented with a pair of edited images for a given original image and editing instruction. They were tasked with selecting which image better fulfilled the instruction or labeling them as being of equivalent quality. This data forms the basis for our pairwise accuracy evaluation. For absolute quality scoring, annotators evaluated individual edited images, assigning both a categorical label (`Good', `Bad', or `Indistinguishable') and an absolute quality score on a scale of 1 to 5. This data was used to analyze the alignment of score distributions between reward methods and human judgment.

\section{Prompt Template}

\begin{promptbox}[Template for Yes/No Evaluation]

\textbf{System Prompt:} You are a helpful assistant.

\textbf{User Prompt:} Here are two images: the original and the edited version. Please evaluate if the edited image successfully meets the following editing instructions and requirements.

Instruction: \{\{instruction\}\}

Requirements: \{\{requirement\}\}
\\ \\
You need to provide a ``Yes'' or ``No'' judgment based on the accuracy and quality of the edit.

Answer ``Yes'' if: The correct object was edited according to the instruction, all requirements were met, and the visual result is high quality.

Answer ``No'' if: The wrong object was edited, the edit fails to meet the requirements, or the visual quality is poor.

\ \

\textbf{Response Format:} 

``Judgement'': [[Yes/No]]

\end{promptbox}

\vspace{13pt}

\begin{promptbox}[Template for Score Evaluation]

\textbf{System Prompt:} You are a helpful assistant.

\textbf{User Prompt:} Here are two images: the original and the edited version. Please evaluate the edited image based on the following editing instructions and requirements.

Instruction: \{\{instruction\}\}

Requirements: \{\{requirement\}\}
\\ \\
You need to rate the editing result from 0 to 5 based on the accuracy and quality of the edit.

0: The wrong object was edited, or the edit completely fails to meet the requirements.

5: The correct object was edited, the requirements were met, and the visual result is high quality.
\\ \\
\textbf{Response Format:}

``Score'': [[rating]](0-5).

\end{promptbox}

\vspace{13pt}

\begin{promptbox}[Template for CoT Score Evaluation]
\textbf{System Prompt:} You are a helpful assistant.

\textbf{User Prompt:} Here are two images: the original and the edited version. Please evaluate the edited image based on the following editing instruction and requirements.

Instruction: \{\{instruction\}\}

Requirements: \{\{requirement\}\}
\\ \\
You need to rate the editing result from 0 to 5 based on the accuracy and quality of the edit. Before that, analyze the image and provide some reasoning process for better evaluation (keep your reasoning concise).

0: The wrong object was edited, or the edit completely fails to meet the requirements.

5: The correct object was edited, the requirements were met, and the visual result is high quality.
\\ \\
\textbf{Response Format:} 

``Reasoning'': ``...'',

``Score'':  [[rating]](0-5).
\end{promptbox}

\end{document}